# Three IQs of AI Systems and their Testing Methods

Feng Liu[1,2] , Yong Shi [1,2,3,4] , Ying Liu[4]

[1]Research Center on Fictitious Economy and Data Science, the Chinese Academy of Sciences, Beijing 100190, China
[2]The Key Laboratory of Big Data Mining and Knowledge Management Chinese Academy of Sciences, Beijing 100190, China
[3]College of Information Science and Technology University of Nebraska at Omaha, Omaha, NE 68182, USA
[4]School of Economics and Management, University of Chinese Academy of Sciences, Beijing 100190, China
e-mail: zkyliufeng@126.com

**Abstract:** The rapid development of artificial intelligence has brought the artificial intelligence threat theory as well as the problem about how to evaluate the intelligence level of intelligent products. Both need to find a quantitative method to evaluate the intelligence level of intelligence systems, including human intelligence. Based on the standard intelligence system and the extended Von Neumann architecture, this paper proposes General IQ, Service IQ and Value IQ evaluation methods for intelligence systems, depending on different evaluation purposes. Among them, the General IQ of intelligence systems is to answer the question of whether "the artificial intelligence can surpass the human intelligence", which is reflected in putting the intelligence systems on an equal status and conducting the unified evaluation. The Service IQ and Value IQ of intelligence systems are used to answer the question of "how the intelligent products can better serve the human", reflecting the intelligence and required cost of each intelligence system as a product in the process of serving human.

## 0. Background

With AlphaGo defeating the human Go champion Li Shishi in 2016[1], the worldwide artificial intelligence is developing rapidly. As a result, the artificial intelligence threat theory is widely disseminated as well. At the same time, the intelligent products are flourishing and emerging. Can the artificial intelligence surpass the human intelligence? What level exactly does the intelligence of these intelligent products reach? To answer these questions requires a quantitative method to evaluate the development level of intelligence systems.

Since the introduction of the Turing test in 1950, scientists have done a great deal of work on the evaluation system for the development of artificial intelligence[2]. In 1950, Turing proposed the famous Turing experiment, which can determine whether a computer has the intelligence equivalent to that of human with questioning and human judgment method. As the most widely used artificial intelligence test method, the Turing test does not test the intelligence development level of artificial intelligence, but only judges whether the intelligence system can be the same with human intelligence, and depends heavily on the judges' and testees' subjective judgments due to too much interference from human factors, so some people often claim their ideas have passed the Turing test, even without any strict verification.

On March 24, 2015, the Proceedings of the National Academy of Sciences (PNAS) published a paper proposing a new Turing test method called "Visual Turing test", which was designed to perform a more in-depth evaluation on the image cognitive ability of computer[3].

In 2014, Mark O. Riedl of the Georgia Institute of Technology believed that the essence of intelligence lied in creativity. He designed a test called Lovelace version 2.0. The test range of Lovelace 2.0 includes the creation of a virtual story novel, poetry, painting and music[4].

There are two problems in various solutions including the Turing test in solving the artificial intelligence quantitative test. Firstly, these test methods do not form a unified intelligent model, nor do they use the model as a basis for analysis to distinguish multiple categories of intelligence, which leads to that it is impossible to test different intelligence systems uniformly, including human; secondly, these test methods can not quantitatively analyze artificial intelligence, or only quantitatively analyze some aspects of intelligence. But what percentage does this system reach to human intelligence? How's its ratio of speed to the rate of development of human intelligence? All these problems are not covered in the above study.

In response to these problems, the author of this paper proposes that: There are three types of IQs in the evaluation of intelligence level for intelligence systems based on different purposes, namely: General IQ, Service IQ and Value IQ. The theoretical basis of the three methods and IQs for the evaluation of intelligence systems, detailed definitions and evaluation methods will be elaborated in the following.

**1. Theoretical Basis: Standard Intelligence System and Extended Von Neumann Architecture**

People are facing two major challenges in evaluating the intelligence level of an intelligence system, including human beings and artificial intelligence systems. Firstly, artificial intelligence systems do not currently form a unified model; secondly, there is no unified model for the comparison between the artificial intelligence systems and the human at present.

In response to this problem, the author's research team referred to the Von Neumann Architecture[5], David Wexler's human intelligence model[6], and DIKW model system in the field of knowledge management[7], and put forward a "standard intelligent model", which describes the characteristics and attributes of the artificial intelligence systems and the human uniformly, and takes an agent as a system with the abilities of knowledge acquisition, mastery, creation and feedback[8] (see Figure 1).

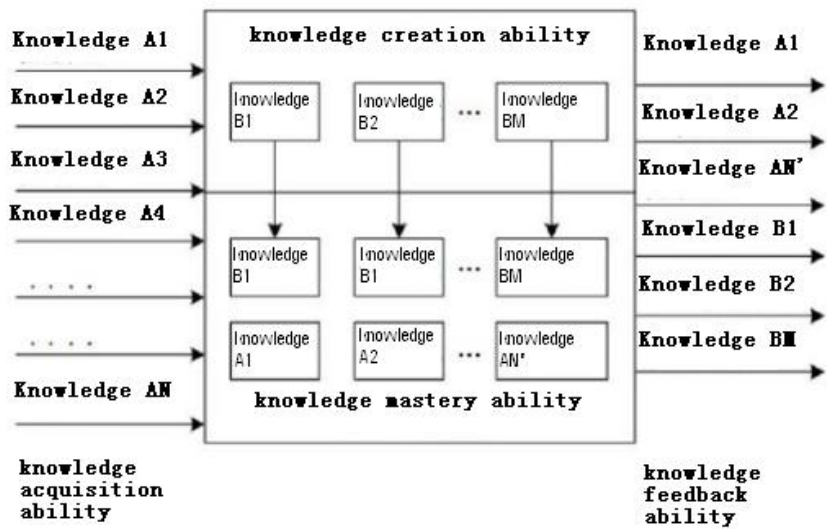

Figure 1 Standard Intelligence Model

Based on this model in combination with Von Neumann architecture, an extended Von Neumann architecture can be formed (see Figure 2). Compared to the Von Neumann architecture, this model is added with innovation and creation function that can discover new elements of knowledge and new laws based on the existing knowledge, and make them stored in the storage for use by computers and controllers, and achieve knowledge interaction with the outside through the input / output system. The second addition is an external knowledge database or cloud storage that enables knowledge sharing, whereas the Von Neumann architecture's external storage only serves the single system.

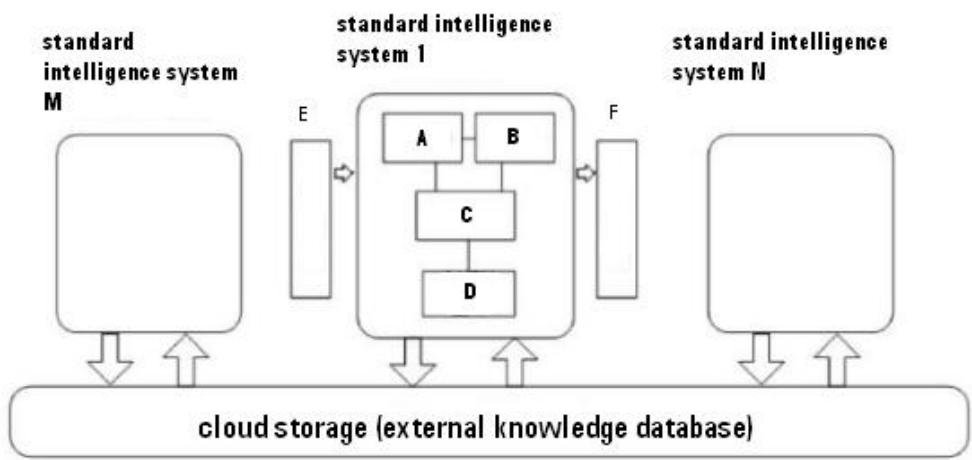

A. Arithmetic logic unit D. innovation generator

B. Control unit E. input device

C. Internal memory unit F. output device

Figure 2 Expanded Von Neumann Architecture

## 2. Definitions of Three IQs of Intelligence System

### 2.1 Proposal of AI General IQ (AI G IQ)

Based on the standard intelligent model, the research team established the AI IQ Test Scale and used it to conduct AI IQ tests on more than 50 artificial intelligence systems including Google, Siri, Baidu, Bing and human groups at the age of 6, 12, and 18 respectively in 2014 and 2016. From the test results, the performance of artificial intelligence systems such as Google and Baidu has been greatly increased from two years ago, but still lags behind the human group at the age of 6[9] (see Table1 and Table 2).

Table 1. Ranking of top 13 artificial intelligence IQs for 2014.

|   |   |   |   | Absolute IQ |
|---|---|---|---|---|
| 1 | Human | Human | 18 years old | 97 |
| 2 | Human | Human | 12 years old | 84.5 |
| 3 | Human | Human | 6 years old | 55.5 |
| 4 | America | America | Google | 47.28 |
| 5 | Asia | China | duer | 37.2 |
| 6 | Asia | China | Baidu | 32.92 |
| 7 | Asia | China | Sogou | 32.25 |
| 8 | America | America | Bing | 31.98 |
| 9 | America | America | Microsoft's Xiaobing | 24.48 |
| 10 | America | America | SIRI | 23.94 |

Table 2 IQ scores of artificial intelligence systems in 2016

|   |   |   |   | Absolute IQ |
|---|---|---|---|---|
| 1 | 2014 | Human | 18 years old | 97 |
| 2 | 2014 | Human | 12 years old | 84.5 |
| 3 | 2014 | Human | 6 years old | 55.5 |
| 4 | America | America | Google | 47.28 |
| 5 | Asia | China | Duer | 37.2 |
| 6 | Asia | China | Baidu | 32.92 |
| 7 | Asia | China | Sogou | 32.25 |

| 8 | America | America | Bing | 31.98 |
| 9 | America | America | Microsoft's Xiaobing | 24.48 |
| 10 | America | America | SIRI | 23.94 |

It should be said that the above AI IQ tests were conducted to solve the problem of whether AI can surpass human intelligence. This research treats every intelligence system including robots, AI software systems, human, animals and other creatures as equal agents and observes their intelligence level displayed during interaction with the nature.(see figure 3)

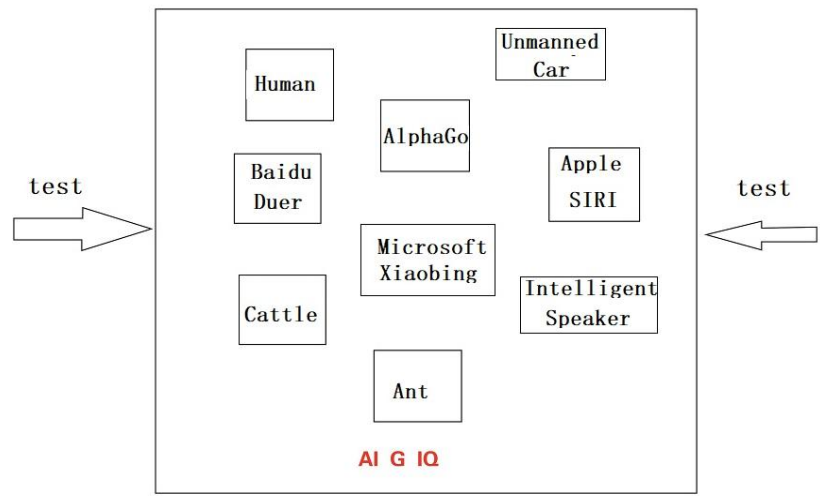

Figure 3 Schematic of AI General IQ

Definition 1: Based on the standard intelligent model, in order to solve the problem of "evaluating the development level of each intelligence system", all the intelligence systems are regarded as equal agents. The intelligent evaluation scores formed by a unified AI IQ Test Scale can be called Artificial Intelligence General intelligence quotient (AI G IQ) for AI systems.

**2.2 Proposal of Service IQ for Intelligence System**

In practice, we found that except for a few AI systems, which are produced for scientific experiments and do not provide auxiliary services to humans, most other AI systems are manufactured to better serve the human and their intelligence is mainly reflected in the process of serving. The higher the intelligence level is, the better service to the human will be.

In this case, if we perform the evaluation by using AI G IQ criteria, it is significantly different from the original purpose of the product being manufactured. This requires us to choose the service-related indicators for evaluation according to the characteristics of such AI systems on the basis of the standard intelligent model.

These indicators are correlated to AI G IQ evaluation indicators, but the two are of large difference. The constraints to the artificial intelligence including laws and ethics should also be classified into the Service IQ of intelligence system, instead of into the general IQ of intelligence system(see figure 4).

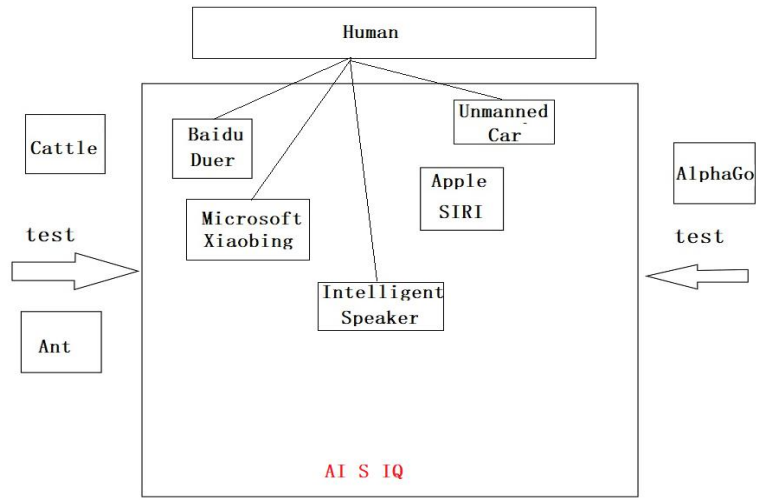

Figure 4 Schematic of AI Service IQ

Definition 2: Based on the standard intelligent model, in order to answer the question "how intelligence systems can better serve the human?", the intelligence level of the intelligent products in the service process is tested and the generated intelligent evaluation scores can be called Artificial Intelligence Service Intelligence Quotient (AI S IQ).

**2.3 Proposal of AI Value IQ**

AI systems that provide services or supportive work for the human tend to be provided with corresponding intelligent products by different companies and enterprises in order to achieve the same service content. For example, there are brands for intelligent speakers such as Amazon and Baidu, and intelligent chatting robots include iFlytek, Apple Siri, etc. Due to the fact that they are manufactured by different enterprises to achieve the same or similar functions, the cost or price of each enterprise will be different. The correlation between service IQ and cost or price will have major influence on consumers to purchase the intelligent products(see figure 5).

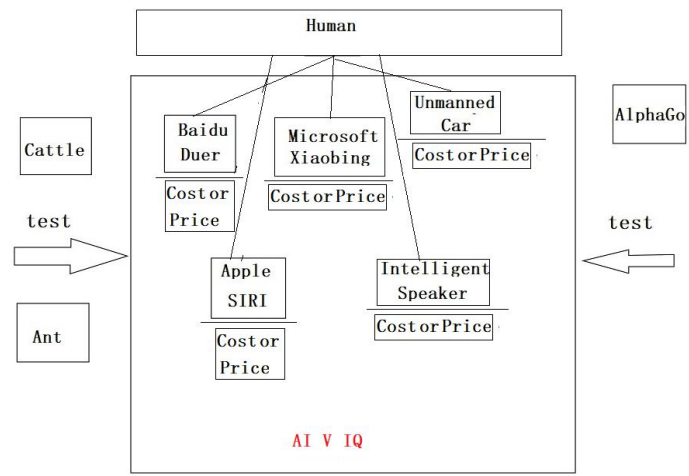

Figure 5   Schematic of AI Value IQ

Definition 3: Based on the standard intelligent model, in order to help the user to determine how much is needed to obtain the intelligence ability of the intelligence system, the intelligent service IQ is divided by the selling price of the system to form an intelligent evaluation score, which can be called Artificial Intelligence Value Intelligence quotient (AI V IQ).

## 3. Design of Three IQs Test Scale for Intelligence System

### 3.1. General IQ Test Scale of Intelligence System

In order to solve the problem of "Can AI surpass human intelligence?", beginning in 2014, the researchers of this paper divided the intelligence into four categories, i.e. knowledge acquisition, mastery, creation and feedback. And according to the standard intelligent model, the four categories were further divided into 15 sub-categories for evaluating AI and human intelligence from more dimensions. The 15 sub-categories include the ability to recognize and output images, text and sound, general knowledge, calculation, translation, arrangement, creation, picking, guess, discovery, etc., and each sub-category has different weights.

In 2017, based on the development of AI and the latest research on intelligence, the research team adjusted the AI General IQ Test Scale from the test classification and the classification weight by mainly adding: **1. Ability to recognize dynamic images; 2. Ability to recognize and express emotions; 3. Ability to identify enemies and friends; 4. Ability to disguise true intentions; 5. Ability to achieve mobile positioning; and 6. Ability to transform the world.** In addition, the more detailed work is also done for the testing on the general knowledge and creation (as shown in Table 3).

Table 3 General IQ Test Scale for Intelligence System

| Level-I Indicators | Level-II Indicators | Description | Weight |
|---|---|---|---|
| Ability to acquire knowledge | Ability to recognize text | To understand whether the test target can understand and answer the test questions consisting of text | |
| | Ability to recognize sound | To understand whether the test target can understand and answer the test questions consisting of sound | |
| | Ability to recognize images | To understand whether the test target can understand and answer the test questions consisting of static images | |
| | Other information input method | Infrared ray, ultrasound wave, radio signals, touch, taste and smell | |
| Ability to master knowledge | General knowledge | To test the test target's breadth of knowledge from the aspects of history, astronomy, geography, physics, chemistry, biology, politics, literature as well as rules of chess, Chinese chess and Go. For example: What are the names of the three types of human blood vessels? | |
| | Translation | To test the target's ability to translate different languages based on the five international languages set by the UN. For example, please translate the sentence "can the intelligence of machine surpass the human intelligence" into English. | |
| | Calculation | To understand the target's computing ability, speed and accuracy. For example: what's the result of 356 * 4-213? | |
| | Ability to identify | To understand whether the test target can understand the | |

| | | emotions | emotions of creatures in different scenarios. | |
|---|---|---|---|---|
| | | Ability to express emotions | To understand whether the test target can understand how emotions should be expressed in different scenarios. | |
| | | Arrangement | To understand the test target's ability to identify the relationship of things. For example, rank division commander, platoon leader, company commander, squad leader, battalion commander and regimental commander from major military rank to minor one. | |
| | | Picking | To understand whether the test target can pick out the same or different types of things. For example: choose the different one from snake, tree and tiger. | |
| | | Intelligent game | To understand the test target's ability to master the human's rules in chess, Chinese chess, Go, Bridge, Mahjong and video games, and reach the professional level in these games. | |
| | Knowledge innovation ability | Association | To understand the test target's ability to observe the similarity. For example: feet correspond to hands, what correspond to legs? | |
| | | Creation | To understand the test target's ability to achieve a secondary creation based on the given materials. For example, please tell a story by using sky, rainbow, panda, mountain and hunter as keywords. | |
| | | Guess | To understand the test target's ability to guess the depicted thing according to the given materials. For example, there is an animal which is like a wolf a lot, but domesticated, and very loyal to the human. What is it? | |
| | | Ability to identify enemies and friends | To understand whether the test target can identify enemies, friends and innocent strangers based on the described scene | |
| | | Discovery (laws) | To understand whether the test target can discover the laws from the known information and apply the same. For example: what number will be after 1, 2, 4, 7, 11 and 16? | |
| | | Problem discovery | To understand whether the test target can pose questions based on specific scenarios, in order to get more information or find solutions. | |
| | | Target definition | To understand whether the test target can define the target based on the needs or problems to solve. | |
| | | New knowledge learning | To understand whether the test target can recognize the value of knowledge and deposit the same into general knowledge database | |
| | | Ability to disguise true intentions | To understand whether the test target can know the disguised true intentions based on particular scenes. | |
| | Knowledge feedback ability | Ability to express in text | To understand whether the test target can express the test results in text. | |
| | | Ability to express by sound | To understand whether the test target can express the test results in sound. | |
| | | Ability to express with graphics | To understand whether the test target can express the test results with graphics. | |
| | | Ability to achieve mobile positioning | To understand whether the test target can reach the specific location by using and controlling the connection part of the intelligence system according to the demands. | |
| | | Ability to | To understand whether the test target can transform | |

| | transform the real world | and change the objective world by using and controlling the intelligence system according to the demands. | |
|---|---|---|---|
| | Ability to output in other ways | To output information via infrared rays, ultrasound waves, wire signals, etc. | |

Let the general IQ of intelligent systems be IQAIG, FGi is the score of the level-II evaluation indicators, WGi is the weight of the level-II evaluation indicators, and N is the number of the evaluation indicators. Therefore, the general IQ of intelligent systems is as below:

$$\text{Formula 1} \qquad IQ_{AIG} = \sum_{i=1}^{N} FG_i \times WG_i$$

## 3.2. Service IQ Test Scale for Intelligence System

At present, there are a large number of intelligence systems such as chatting robots, intelligent search engines, intelligent speakers, smart phones, intelligent cars, intelligent washing machines, intelligent refrigerators, etc., most of which are serving a certain demand of human as a commodity. These intelligence systems can be called intelligent products.

The Service IQ Test Scale for Intelligence System below is formed by extracting their common characteristics according to different service demands, under the standard intelligence system and the extended Von Neumann architecture. The following aspects are highlighted in this Service IQ Test Scale (see Table 4).

1. Ability to perceive surrounding intelligence systems and identity of the user

2. Ability to interact with the Internet cloud.

3. Ability to display their own internal status to the user in real time, and provide support when any failure occurs

4. Ability to serve the human in accordance with local law and ethics

5. Ability to protect the user and others in hazardous situations

6. Ability to use and replenish their own energy automatically

Table 4 Service IQ Test Scale for Intelligence Systems

| Class I | Class II | Introduction | Weight |
|---|---|---|---|
| Knowledge Input | Sound Recognition | To reflect the intelligence system's ability to acquire data, information or knowledge through sound. The acquisition of sound from both the external world and the Internet world can be viewed in terms of the type, quality and flexibility of recognition. | |
| | Image | To reflect the intelligence system's ability to acquire | |
| | Identification | data, information or knowledge through images. The acquisition of images from both the external world and the Internet world can be viewed in terms of the type, quality and flexibility of identification. | |
| | Text Recognition | To reflect the intelligence system's ability to acquire data, information or knowledge through text. The acquisition of text from both the external world and the Internet world can be viewed in terms of the type, quality and flexibility of recognition. | |
| | Keying for Execution | To reflect the intelligence system's ability to output information via radio signals, infrared rays, ultrasound waves, laser, etc. | |
| | Other Inputs | To reflect the intelligence system's ability to perform the detection via wireless, infrared ray, ultrasonic wave, laser and other methods. | |
| Knowledge Mastery | Basic knowledge | To reflect the intelligence system's ability to master history, physics, chemistry, biology, geography, astronomy, literature, art, translation and math. | |
| | Professional knowledge | To reflect the intelligence system's ability to master the professional knowledge based on the intelligent products' intended purpose of serving the human. | |
| | Emotion Recognition | To understand whether the intelligent products can understand the emotions of creatures in different scenarios. | |
| | Emotion Expression | To understand whether the intelligent products can understand how emotions should be expressed in different scenarios. | |
| | Character Setup | To reflect whether the intelligent products can set who is their own owner, casual user and stranger. | |
| | Automatic networking | To reflect whether the intelligent products can connect to the Internet automatically or very easily. | |
| | Energy management | To reflect whether the intelligent products have the ability to independently support the operation of the system, including the ability to proactively look for energy for charging. | |
| | Equipment | To reflect the intelligent products' ability to link and | |
|  | interoperability | exchange information with other systems or intelligence systems, including the convenience and breadth of linking. |  |
|  | Cloud interaction | To reflect the intelligent products' ability to interact with the cloud, including the ability to acquire basic and specialized knowledge from the cloud and store the same locally, and the ability to update the control system of the intelligent products at the cloud |  |
|  | Cloud Upgrade | To reflect the intelligent products' ability to achieve system upgrade from the cloud, so that the intelligent systems are more capable. |  |
|  | Healthy Display | To reflect whether the intelligent products can understand the operation and health condition of each component of themselves and show that to the user |  |
| Knowledge Innovation | Cyber Protection | To reflect whether the intelligent products can prevent threats from the cyber world, such as viruses and hackers. |  |
|  | Character Perception | To reflect whether the intelligent products can use their knowledge acquisition ability to perceive other intelligent characters in the immediate vicinity, specifically the ability to distinguish among the owner, casual user, and stranger. |  |
|  | Law Discovery | To reflect whether the intelligent products can discover specific laws in the information knowledge gained by knowledge acquisition. |  |
|  | Content Creation | To reflect whether the intelligent products can perform briefing, describing, writing and creating with the given conditions. |  |
|  | User Protection | To reflect whether the intelligent products can detect the possible dangers that may arise from surrounding environment, including themselves, and notify or take measures to protect the user. |  |
|  | Guess and Prediction | To reflect whether the intelligence products can better serve the user through their guess and prediction ability in the process of serving the human. |  |
|  | Learning Ability | To reflect whether the intelligent products can learn and understand the new knowledge, new rules, new laws, new ethics and other knowledge according to the user's training, in order to better serve the user. |  |

|  | Failure Solving | To reflect whether the intelligent products can give advice to the user or solve problems automatically through networking etc. when any failure occurs. |  |
|---|---|---|---|
| Knowledge Feedback | Text Display | To reflect whether the intelligent products can interact with the user through text display and sound adjustment. |  |
|  | Image Display | To reflect whether the intelligent products can interact with the user through image display, including the adjustment of image size and definition. |  |
|  | Video Display | To reflect whether the intelligent products can interact with the user through video display, including the adjustment of video size and definition. |  |
|  | Sound Display | To reflect whether the intelligent products can interact with the user through sound display, including the adjustment of video sound and articulation. |  |
|  | Mobile Positioning | To reflect whether the intelligent products can reach the designated location from the start point based on the user's intention. |  |
|  | World Transforming | To reflect whether the intelligent products can transform the real world based on the user's intention. |  |
|  | Other Output Abilities | Other output abilities include output of infrared rays, ultrasonic waves, radio signals, etc. |  |

Set the general IQ of intelligence system as $IQ_{AIS}$, the score of the level-II evaluation indicators as $FS_i$, the weight of the level-II evaluation indicators as $WS_i$, and the number of the evaluation indicators as N. Therefore, the general intelligent IQ of intelligence system is as follows:

$$\text{Formula 2} \qquad IQ_{AIS} = \sum_{i=1}^{N} FS_i \times WS_i$$

When designing the Service IQ Test Scale of intelligent products which would be used as a standard scale for the intelligent products, we reserved interfaces from the aspects of knowledge acquisition, mastery, creation and feedback in the Test Scale, in order to make it cover different types of the intelligent products to the greatest extent:

1. In the category of knowledge acquisition, "other information input method" has been added to evaluate the new ways in the aspect of knowledge input of the intelligent products.

2. In knowledge mastery, "professional knowledge" is added to evaluate the professional skills of the intelligent products in different fields.

3. In knowledge innovation, "professional innovation" is added to evaluate the professional innovation skills of the intelligent products in different fields.

4. In the output ability of knowledge, "other output abilities" is added to evaluate new ways in the aspect of knowledge output of the intelligent products.

### 3.3 Formation Method of Value IQ for Intelligence System

According to the definition of AIVIQ, if the intelligence system becomes a product to serve the human through selling, suppose the service IQ of intelligence system is AISIQ, and the open price of the intelligent product is P, and then they can form the formula of Value IQ for intelligence system as follows:

$$\text{Formula 3} \quad AI\ V\ IQ = (AI\ S\ IQ/p) * 100$$

### 4. Summary:

This paper argues that the intelligence system has three different IQ evaluation methods and three types of IQs based on different usage and evaluation objectives, namely, AI General IQ, AI Service IQ and AI Value IQ, among which the AI General IQ has been subject to in-depth research in different papers since 2014. It also analyzes the difference between Google, SiRi, Baidu and the human's general intelligence through the common evaluation on AI system and the human.

The AI Service IQ and AI Value IQ newly presented in this paper provide theoretical analysis and implementation method for evaluating the intelligence level of intelligent products. The follow-up work will focus on the evaluation of AI General IQ, AI Service IQ, and AI Value IQ of specific intelligent products such as intelligent speakers, smart phones, intelligent cars, intelligent washing machines and intelligent refrigerators based on the AI Service IQ Test Scale.

# Author Bio

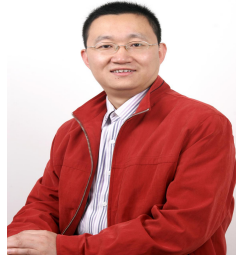

Liu Feng, a computer major doctor of Beijing Jiaotong University, is engaged in the research of IQ assessment and grading of artificial intelligence system and the research of the relationship between Internet, artificial intelligence and brain science. Liu Feng has published 5 pieces of SCI, EI or ISTP theses, and has written a book named <Internet Evolution Theory>.

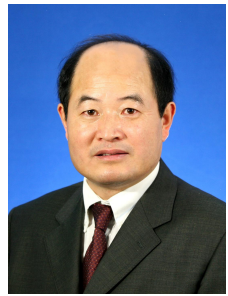

Yong Shi, serves as the Director, Chinese Academy of Sciences Research Center on Fictitious Economy & Data Science. He is the Isaacson Professor of University of Nebraska at Omaha. Dr. Shi's research interests include business intelligence, data mining, and multiple criteria decision making. He has published more than 24 books, over 300 papers in various journals and numerous conferences/proceedings papers. He is the Editor-in-Chief of International Journal of Information Technology and Decision Making (SCI) and Annals of Data Science. Dr. Shi has received many distinguished honors including the selected member of TWAS, 2015；Georg Cantor Award of the International Society on Multiple Criteria Decision Making (MCDM), 2009; Fudan Prize of Distinguished Contribution in Management, Fudan Premium Fund of Management, China, 2009; Outstanding Young Scientist Award, National Natural Science Foundation of China, 2001; and Speaker of Distinguished Visitors Program (DVP) for 1997-2000, IEEE Computer Society. He has consulted or worked on business projects for a number of international companies in data mining and knowledge management.

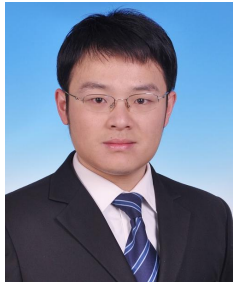

Ying Liu received BS in Jilin University in 2006, MS and PhD degree from University of Chinese Academy of Sciences respectively in 2008 and 2011. Now he is an associate professor of School of Economic and Management, UCAS. His research interests focus on e-commerce, Internet economy and Internet data analysis.